# Motion Planning for Autonomous Ground Vehicles Using Artificial Potential Fields: A Review

Aziz ur Rehman[a, *], Ahsan Tanveer[a, *], M. Touseef Ashraf[a], Umer Khan[a]

[a] *Department of Mechanical and Aerospace Engineering, Institute of Avionics and Aeronautics, Air University, Islamabad 44000, Pakistan*

* Corresponding author: Aziz ur Rehman; Ahsan Tanveer, Email: azizkhokhar11@gmail.com; ahsan.tanveer@mail.au.edu.pk

A B S T R A C T

Autonomous ground vehicle systems have found extensive potential and practical applications in the modern world. The development of an autonomous ground vehicle poses a significant challenge, particularly in identifying the best path plan, based on defined performance metrics such as safety margin, shortest time, and energy consumption. Various techniques for motion planning have been proposed by researchers, one of which is the use of artificial potential fields. Several authors in the past two decades have proposed various modified versions of the artificial potential field algorithms. The variations of traditional APF approach have given answer to prior shortcomings. This gives potential rise to a strategic survey on the improved versions of this algorithm. This study presents a review of motion planning for autonomous ground vehicles using artificial potential fields. Each article is evaluated based on criteria that involve, the environment type, which may be either static or dynamic, the evaluation scenario, which may be real-time or simulated, and the method used for improving the search performance of the algorithm. All the customized designs of planning models are analysed and evaluated. At the end, the results of the review are discussed and future works are proposed.

## 1. Introduction

Autonomous Ground Vehicles or AGVs are autonomously navigated vehicles that rely on advanced sensors, cameras, and software tools to navigate and operate independently. AGVs are land-based counterparts to autonomous aerial vehicles (AAVs), [1] and autonomous underwater vehicles (AUVs) [2]. A variety of dull, dirty, and dangerous activities are nowadays being performed through the use of AGV robots. Their applications can be found in agriculture and pest control, security and surveillance, exploration and navigation, reconnaissance, and scouting operations. The integration of autonomy in AGVs make them capable of making independent decisions and taking actions based on the surrounding environment information. The simplified system layer of an autonomous ground vehicle system is shown in Figure 1.

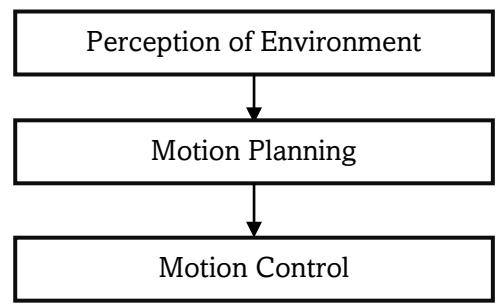

Figure 1. Layers of an Autonomous System.

The optimized navigation of an AGV, while avoiding obstacles in the static and dynamic environment, requires the use of optimized path and motion planning algorithms. The algorithm determines a feasible path from a start position to a goal (target) position and optimize the plan along some defined criteria. These criteria or performance measures of the algorithm include time and distance minimization, safety, comfort, and controlling effort, which are used to help identify the best feasible path.

Motion planning of UGVs is a challenging task. it is hard to solve it in a single integrated process. For this reason, motion planning is solved by dividing the whole problem into multiple layers of abstraction. At the top-level layer, the planner handles long-term planning. In this layer, the planner defines the problem over the entire map of the driving scenario, that is from the current location in the environment, through the road network to the target location. This can be termed global path planning. Next is the behavioural planner of the vehicle and is responsible for safe behavioural actions and execution of manoeuvres along the path given by the top-level layer. These behavioural actions may include lane changing, lane keeping, parking, or intersection crossing. At the end layer, the local planning is executed. This layer performs immediate or reactive planning to dynamic obstacles discovered along the global path, as well as define the motion and velocity profile, that is smooth, safe, efficient, and satisfies the constraints of the vehicle and the environment.

Path planning can be divided into static path planning and dynamic path planning from the perspective of whether the environment obstacles are moving or not. It is noted that for global path planning, the environment is generally assumed to be static. This layer of planning is performed before the robots begin to navigate hence offline, while that is not the case for local planning. During the local planning phase, the planner executes reactive planning and hence online, while traversing on the global path based on the data from the on-board sensors. This assumes incomplete knowledge of the environment, hence optimal for navigation in dynamic environments. Figure 2 best illustrates the scenario for the local (online) and global (offline) planning for the AGV robot.

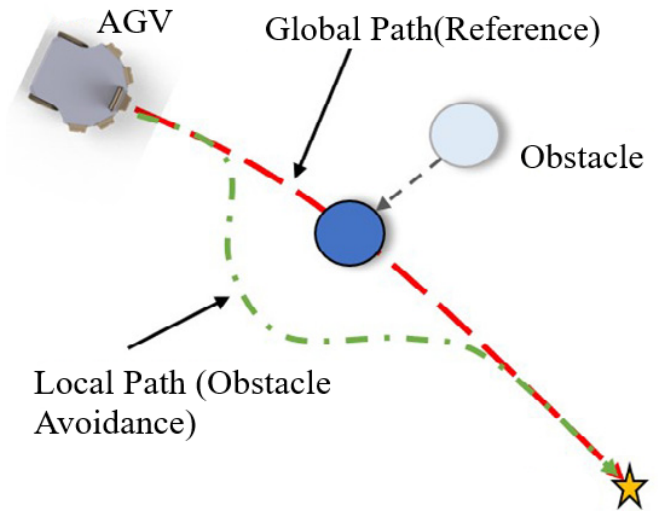

Figure 2. Scenario for Local and Global Path Planning [3].

The advancements in the development of autonomy in AGVs have led to highly accurate 3-D environmental models. However, for practical application, the autonomous vehicle has to navigate a safe and optimized path plan. It is hard to solve the path planning problem, which is generally considered a non-deterministic polynomial-time ("NP") problem, considering more complications, such as the increase in the degree of freedom (DOF) of the system in a 3-D environment. Additionally, the problem formulation of the AGV system involves constraints like non-holonomic motion constraints or dynamic constraints, which increase the complexity of the problem [3]. In other cases, the computational efficiency and the compatibility of the motion plan with the assumptions required for optimal control, to make the path traceable are also a significant challenge [4].

A variety of different surveys have been presented, such as in [3], a systematic survey of path planning was presented, which classified a variety of planning techniques as classical and heuristic techniques. In [4], a survey of planning and control was presented, giving attention to tailoring and integration the two designs. While, different papers have classified the planning techniques according to the nature of their review, such as traditional and machine learning algorithms [5]. A more complete classification of path and motion planners was presented in [6]. Going deeper into the study, there have been a growing number of surveys on motion planners. However, a review of modified and improved versions of one particular algorithm are undiscovered. Limited research has been found, such as Short et al. [7] gave a review of sampling-based planners. Similarly, just recently in 2023, Paliwal [8] published a survey of improved versions of the A-star family of motion planners, which are similar in standings to the artificial potential field algorithms.

In this review, we have selected to review versions of artificial potential field (APF) as the planning algorithm for autonomous ground vehicles. There have been a variety of different improved versions of APF planning algorithm. Such variations in the traditional APF algorithms have given answer to prior shortcomings. Ultimately, this gives potential importance to a strategic survey on versions of this algorithm. Every article in this survey is evaluated based on criteria that involve, the environment type, which may be either static or dynamic, the evaluation scenario, which may be real-time or simulated, and the method used for improving the search performance and optimization of the APF path plan. This paper presents a review that lists different variants of the APF motion planning approach and discusses results.

The paper is organized as follows: Section 2 provides a comprehensive overview of the structure of the APF algorithm; Section 3 will cover review and analysis of the different variants of the APF motion planning for AGVs; In Section 4, results of the literature will be discussed; At the end, in section 5, a conclusion and future recommendations will be drawn out from the paper.

## 2. Artificial Potential Field

Artificial potential field, abbreviated as APF was first proposed by Khatib et al. [9]. It is a well-recognized grid-based motion planning approach that is used nowadays as a reactive planner for obstacle avoidance. It can be used for planning in both offline and online maps, in other words, for both global and local planning.

The basis of the APF method in motion planning follows the natural characteristic of an electrostatic potential. The interaction of the electrostatic particles in physics is given as

$$F = \frac{-kq_1q_2}{r^2}$$

where $k$ is the interactions constant, $q_1$ and $q_2$ are the electric charges of the particles, and $r$ is the distance between these particles. If the charge of the two particles is the same, it repels, Conversely, if the charges are opposite, they attract. Such is the formulation of the APF problem. Both, the goal point and the obstacles, and in some cases the road boundaries exert force fields on the ego vehicle. The goal exerts a force of attraction $F_{att}$ while the obstacle and road boundaries exert a repulsive force $F_{rep}$. The attraction force is also termed as the gravity force. This means that the traditional APF is a combination of two different potential functions. If we consider the position of a robot at a particular time (t) to be $q = [x\ y]^T$, then the force of attraction $F_{att}$ can be modelled as

$$F_{att}(q) = -\nabla V_{att}(q)$$

where

$$V_{att}(q) = \frac{1}{2}\xi p^2(q, q_{goal})$$

Here $V_{att}(q)$ is the attractive potential, $\xi$ is the gain parameter which has a position value, while, $p(q, q_{goal})$ is the distance between the position of the ego vehicle q and the goal location $q_{goal}$. Mathematically $p(q, q_{goal}) = ||q_{goal} - q||$, which defines that the value converges to zero as the ego vehicle moves towards the target location.

Similarly, the repulsive force of attraction $F_{rep}$, which is from the obstacle is modeled as

$$F_{rep}(q) = -\nabla V_{rep}(q)$$

Where,

$$V_{rep}(q) = \begin{cases} \dfrac{\eta}{\sqrt{\rho(q, q_{abs})}} & if\ \rho(q, q_{abs}) \leq \rho_o \\ 0 & if\ \rho(q, q_{abs}) \geq \rho_o \end{cases}$$

Here $\eta$ is the repulsive gain parameter, $V_{rep}(q)$ is the repulsive potential, while $\rho(q, q_{abs})$ is the distance between the obstacle and the ego vehicle position. The c-obstacle of the ego vehicle dimension is represented by the variable $\rho_o$ which is a positive number. The resultant force of the attractive and repulsive forces is represented as,

$$F_{total} = F_{att} + F_{rep}$$

This resultant force $F_{total}$ is a vector quantity, which represents the direction of movement of the ego vehicle [10]. An illustration of this resultant force is shown in Figure 3. The resultant force provides the reference for the AGV local path generation and control. The deployment of electrostatic potential for path generation can also serve as a reactive planner for obstacle avoidance, hence suitable for local path planning in undiscovered surroundings.

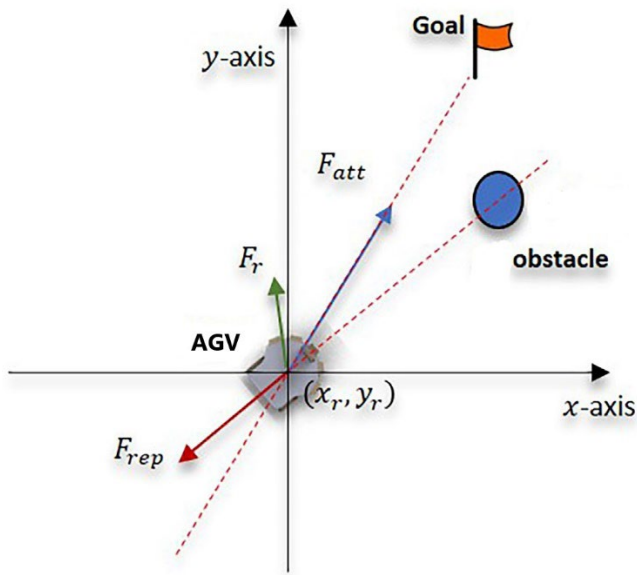

Figure 3. Illustration of the resultant force that is generated on the AGV [3].

*2.1 Shortcomings of unmodified Traditional APFs in AGV motion planning*

The use of artificial potential fields for local path planning has received much attention because of its simplicity and low computation cost for obstacle avoidance in real time. However, as this technique was applied in complex environments with dense obstacles and in real time experiments, it covered some shortcomings.

One of the drawbacks of the APF method in motion planning is the problem of local minima. This is due to the robot detecting its current location as the goal point because the attractive potential and repulsive potential have become equal. The AGV is then trapped before reaching the goal location, which is the global minima. The problem of local minima is inevitable because the robot has only local information of its current surroundings, and the issue can only be detected if the robot has information of the whole environment. The planner also has navigation issues and falls into traps for narrow passages in environments.

Autonomous ground vehicles are non-holonomic, however, the APF assumes the robot to be point mass. It means that the APF gives the resulting forces which may be in any direction, rejecting the possible constrained movement of the robot. Another inherent issue of the classical APF is the goal non-reachable with obstacles nearby (GNRON) problem In some cases, Oscillations also occur in path generation, which are not acceptable to be neglected in real-time [10].

The solution to these drawbacks has been an active area of research for the past decades. Different researchers have proposed different improvements in the traditional APF method for motion planning and real-time obstacle avoidance. Our study gives a review of the different solutions to the shortcomings associated and gives a literature on improved and modified APF motion planning models.

### 3. Literature Review

There have been a variety of methods to improve the performance of traditional artificial potential field motion planning with real-time obstacle avoidance. Some of these methods include modification in the repulsive or attractive functions, and addition of some other potential fields such as road boundary repulsive fields, or escaping potential functions. Other methods involve the inclusion of a virtual obstacle or goal that provides the force required to escape the local minima and solution of the GNRON problem. A variety of methods also involve the use of APF combined with other planning algorithms. In this strategic review of the improved artificial potential field algorithms, 32 articles were reviewed. The aims of the review were to analyse the various methods employed to enhance the APF motion planner, as well as its integration with other techniques. The assessment also has considered environmental conditions (static or dynamic) and the specific scenarios of experimentation or simulation. A brief explanation of the articles reviewed is below

A solution for the local minima problem was presented using additional components and external forces fields in the repulsive potential field function of the APF motion planner [10, 11]. The modified APF was then able to overcome the local minima and proceed towards the global minima, which represented the goal location. In [12], the formulation of attractive and repulsive functions in APF were modified for planning. In [13, 14], the issue of the local minima was solved by an appropriate steering angle, and by setting a step function for jumping out of the local minima. In [15], for collision avoidance of AV, an improved APF was proposed with the modification of the road potential field, attractive and repulsive fields. In addition to these modifications, the influence of obstacles on the collision areas was redesigned and safety factors were introduced. In [16], a boundary repulsion potential field was introduced for limiting the range of vehicle motion. Moreover, optimization of the repulsive potential function, and distance factor between the target and the vehicle was also introduced for GNRON issues.

Duan et al. [17], suggested an improved APF which utilized a virtual escaping potential field function. This function served as a trigger and influencing factor for situations where the ego vehicle gets stuck and falls into the local minimal location. In [18], a potential function that used the distance between the robot and the goal point for solving the issues related to the local minimum was proposed. In [14], a safety distance and correction factor was introduced for solving the GNRON problem.

Also, Szczepanski et al. [19] proposed the use of augmented reality for detecting the local minimum during navigation. The proposed planner detects if there is a local minimum in the plan currently traversing through the LIDAR. In case there is a local minimum, the planner activates a trigger to generate a virtual wall and initiate a bypassing procedure. Haoyang Li [20], on the other hand used LIDAR to obtain the distance of the ego vehicle from the target location, which he used to identify whether the planner was stuck in a local minima. If the planner is stuck, his modified APF initiates the correction procedure for the robot to separate itself from the equilibrium state.

In [21], improved APF method was proposed in response to the shortcomings faced in traditional APF methods, where different terms were added, which included a distance correction factor, repulsive fields of the dynamic road environments, the velocity, and the accelerations of the vehicle. The invasive weed algorithm was also then integrated to further improve APF motion planning and obstacle avoidance.

Zhang et al. [22] proposed an APF planner, where the repulsive function was modified and he also introduced the selection of a new virtual goal point in case the planning gets stuck. In [23], Szczepanski et al. proposed to place an additional virtual obstacle called top quark in critical areas of the environment, which provided a supplementary repulsive force at critical areas of the environment for energy-efficient APF motion planning. In addition, while considering the stagnation-free path of the planner, a temporary goal point was also selected. This approach was very useful in allowing the planner to reduce the travel length, and the traversing time, increasing the smoothness of the trajectory, and evasion of the local minima. H. Wang et al. [24] suggested a modified APF approach for local minimum problem in patrol swarms. He proposed the adjustment of weights of the attractive and repulsive forces to ensure the local minimum trap evasion.

Similarly, Songtao Xie et al. [25] proposed modified APF methods for distributed motion planning, the velocity difference potential field (VDPF) and acceleration difference potential field (ADPF). The authors used multiple field functions for the algorithm which include the attractive quadratic potential field between the ego vehicle and the virtual goal, the repulsive potential field from the human-operated vehicles, the repulsive potential field from the automated vehicles, and finally the repulsive potential field generated from to road boundaries.

In [13], a road boundary potential field function was added to a traditional APF for improved intelligent vehicle motion planning with obstacle avoidance. Furthermore, in this topic of research, Lazarowska [26] proposed a discrete APF planning method, where the path planning of a differentially driven mobile robot was generated by taking into account different parameters which include the dynamic obstacles, path length, and run time of the problem.

Hongcai Li et al. [27] proposed a DynEFWA (dynamic enhanced firework algorithm) APF approach for motion planning. He introduced the modification of the repulsive and attractive functions on the basis of the shapes of obstacle and the driving characteristics constraints of the vehicle, such as the brake distances and lateral distances of the external vehicles. The authors proposed to use fireworks algorithm for further cost optimization of the DynEFWA-APF path planned.

In addition to the APF algorithm, the A-star algorithm is a graph traversal and path search algorithm that is widely due to shortest optimal path generation and performance efficiency. The A-star algorithm uses heuristics functions guide its search and find the path between multiple nodes on a graph with the smallest cost. Although the A-star algorithm obtains a relatively short path, it is unable to handle dynamic characteristics of environment, the artificial potential field method however can handle dynamic obstacles but the generated path is much longer than the A-star algorithm. Hence in [28, 29], a fusion of A-star algorithm with APF was proposed for optimized motion planning. Similarly, Guodong Du et al. [30] introduced a hierarchal approach to obtain optimal motion planning. He proposed the incorporation of APF with the global reference path that was generated using modified A-star and weighted regression for optimal sequences of motion planning.

In addition to the heuristic nature of the A-star algorithm, bacteria evolutionary algorithm (BEA) is also a nature inspired approach, based on the evolutionary process of bacterial gene recombination. In [31], utilizing the advantages and trimming down disadvantages in traditional APFs, bacterial evolutionary algorithm was proposed for combination for flexible path planning.

Probabilistic roadmap (PRM), which is a graph search-based algorithms is useful for finding best possible paths in grid maps. PRM objects and generates nodes in the free and occupied spaces of the environment and creates a connection between them. While, cost functions are also designed that help in identifying the best possible path for obstacle avoidance. In [32], APF was used for improving the locations of the PRM nodes, while the A-star algorithms was used for the finding the shortest plan.

Similar to the bacterial evolutionary algorithm, chaotic optimization algorithm is also an optimization algorithm, which uses chaotic maps, influenced by the properties of ergodicity and randomness, to enhance search optimization and performance. In [33], a combination of chaotic optimization algorithm with the APF for optimal planning was proposed for motion planning. This approach also introduced adaptation of potential function as objective function and proposed use of robot direction of movement as the control variable.

In [34], an adaptive approach to artificial potential field combined with ant colony optimization for motion planning and optimal obstacle evasion was considered based on the relative position and velocity of the robot to the obstacles.

Genetic algorithm is also a heuristic algorithm, which is based on the Darwinians theory of natural evolution. This algorithm was proposed by Feng Liu et al. [35] for optimisation of the gravity and repulsive functions in the APF algorithm. The optimization algorithm used cost functions to find the lowest possible potential energy location, which is at the goal location, and for determination of the step size and movement of direction for the robot. He also used virtual target points for escaping local minima.

Particle swarm optimization algorithm, which is inspired by the collective behaviour of living organisms, such as birds or fishes, was introduced for obtaining optimum values of APF factors in [36]. This approach iteratively enhanced the APF until the shortest path was obtained, which was then smoothened using the spline equation.

Rapidly exploring random trees is sampling based motion planning approach that searches through non-convex dimensional spaces by randomly building a space-filling tree. In [37], a hybrid motion planning algorithm was proposed utilizing a combination of informed rapidly exploring random trees-star (informed-RRT*) with APF.

Chen Zheyi and Xu Bing [38] suggested using fuzzy control method for improving the adaptability of APF path planning. Additionally, by adding an angle repulsive function to increase the repulsion as the distance between the target point and AGV increases and in other cases a dragging force effect as the distance decreases, for counteracting strong gravitational forces. In [39], the fuzzy algorithm was proposed as an alternative planner for a vehicle, when in local minimum location in the APF path plan. The planner used fuzzy algorithm when in local minimum, and switched back to the APF when the robot moved away from that location.

Simulated annealing is a probabilistic method for path planning, which was proposed in combination with APF for solving problem of local minima and oscillations in the motion plan [40]. Additionally, an escape module was developed for local minimum with a modified resultant force for optimal global planning solution.

In [41], an innovative approach to APF motion planning was proposed. The proposed APF method utilized machine learning for training a model according to different driver characteristics. This model was then used for designing different repulsive field functions according to their respective driver characteristics and road environments. The resulting planner demonstrated adaptability to different surrounding vehicle environments and showed behaviour akin to that of a human-like driver.

## 4. Discussion of the literature

A strategy review of methods used for solving the problems associated with APF motion planners was presented. In this strategic review of the improved artificial potential field algorithms, 32 articles were reviewed. It was observed that although a number of articles [10, 11, 16, 22-25] have proposed modifications in the repulsive and attractive functions, added additional boundary or repulsive functions, used virtual obstacle and goal points, these approaches are not applicable in dynamic or real-world experiments. It was also observed that the majority of the articles have based their research on simulated scenarios, without real-world implementations [10, 11, 17, 18, 20, 22, 25, 41]. A number of authors focused on solving the problems related to local minima and GNRON [10, 18], where they have assumed static obstacles. However, these issues can also arise when the AGV traverses in real-time with dynamic obstacles. Although the APF has solved a variety of issues, Triharminto et al. [10] discussed that oscillations were still occurring in the path plan. Due to this reason, [27, 29, 31, 36] proposed APF combined with other planning algorithms and obtained good results [42].

## 5. Conclusion

Autonomous systems are an increasing area of research and hold significant promise. An autonomous ground vehicle system is a unification of different architectural designs, one of which is motion planning. This paper presents a strategic review of the modification and improvements involved in autonomous motion planning using artificial potential fields. Different approaches were reviewed for solving the motion planning issues, which involve the use of additional repulsive functions, correction factors, virtual obstacle and goal points, and combination with other algorithms. Several articles were found that had designed modified APF integrated with optimal control designs, where the vehicles were modelled according to the kinematic and dynamic constraints of the vehicle for obstacle avoidance. However, they were neglected in this research as those articles were more in correlation with optimal control. Hence, this can be proposed as future work.

**Appendix A**

| Sr. No. | Title | Modification to APF | Environment (static or dynamic) | Evaluation (simulation or Realtime) |
|---|---|---|---|---|
| 1 | A novel of repulsive function on artificial potential field for robot path planning | Modified the repulsive force | Static | Simulation |
| 2 | Active obstacle avoidance method of autonomous vehicle based on improved artificial potential field | Added a virtual escaping force | Dynamic | Simulation |
| 3 | Distributed motion planning for safe autonomous vehicle overtaking via artificial potential field | Modified the resultant force | Dynamic | Simulation |
| 4 | Autonomous vehicle path planning based on driver characteristics identification and improved artificial potential field | Modified the repulsive field function to a ML trained driver characteristics model | Dynamic | Simulation |
| 5 | Efficient local path planning algorithm using artificial potential field supported by augmented reality | Used lidar data to detect local minima and modified the APF to bypass it. | Static | Simulation and real time |
| 6 | Energy efficient local path planning algorithm based on predictive artificial potential field | Added virtual temporary obstacles and goal points at critical areas | Static | Real time |
| 7 | Robotic path planning strategy based on improved artificial potential field | Used lidar to sense the distance b/w the vehicle and the target, which helped in identifying whether the robot is trapped or not and perform evasion | Static | Simulation |
| 8 | Trajectory generation and tracking control of an autonomous vehicle based on artificial potential field and optimized backstepping controller | Modified the repulsive force | Static | Simulation |
| 9 | Path planning based on improved artificial potential field method | Modified the repulsive force and added a virtual goal in case of a local minima trap | Static | Simulation |
| 10 | A new method for robot path planning based artificial potential field | Used potential field filling for GNRON problem and regression search for optimization of path | Static | Simulation |
| 11 | Research on active obstacle avoidance of intelligent vehicles based on improved artificial potential field method | Modified the repulsive force, and added a step function for jumping out of the local minima trap | Static and dynamic | Simulations and real time |
| 12 | Hybrid ant colony and immune network algorithm based on improved APF for optimal motion planning | Combined the use of APF with ant colony network algorithm | Static and dynamic | Simulations and real time |
| 13 | APF-IRRT*: an improved informed rapidly-exploring random trees-star algorithm by introducing artificial potential field method for mobile robot path planning | Combined the use of APF with improved rapidly exploring random trees-star algorithm | Static and dynamic | Simulations and real time |
| 14 | Unmanned vehicle route planning based on improved artificial potential field method | Added steering angle for solving local minima problem. Also added a safety distance and a correction factor for GNRON problem | Static | Simulation |
| 15 | Path planning using artificial potential field method and a-star fusion algorithm | Fusion of A-star and APF algorithms | Static and dynamic | Simulation |
| 16 | Discrete artificial potential field approach to mobile robot path planning | Customized APF approach to a 2D discrete configuration space. Additionally, used Path Optimization algorithm to improve smoothness and length of path | Static and dynamic | Simulation |
| 17 | An optimization-based path planning approach for autonomous vehicles using the DynEFWA-artificial potential field | Constraints of vehicle dynamics, driver characteristics and shapes of obstacles were taken into consideration. Also, used fireworks algorithm for path cost optimization | Static and dynamic | Simulation |
| 18 | Collision avoidance method of autonomous vehicle based on improved artificial potential field algorithm | Introduced safety distance and road potential field in the modified APF and redesigned the influence range of obstacles based on the collision areas and corresponding safety distance | Static and dynamic | Simulation |
| 19 | Research on automatic driving trajectory planning and tracking control based on improvement of the artificial potential field method | Added distance correction factor, with the addition of dynamic road repulsive field, velocity repulsive field, and acceleration repulsive field. Used invasive weed algorithm for improved motion planning | Static and dynamic | Simulation |
| 20 | Dynamic path planning of mobile robot based on artificial potential field | APF combined with simulated annealing algorithm to solve local minimum and oscillation problems, additionally an escape module with modified resultant force was also designed | Static and dynamic | Simulation |
| 21 | Research on intelligent vehicle path planning based on improved artificial potential field method | Added distance correction factor. Also added the use of boundary repulsive potential field, and further optimized the repulsive potential field of obstacle's | Static and dynamic | Simulation |
| 22 | AGV path planning based on improved artificial potential field method | Angle function was added to classical APF, while fuzzy control idea for used for further enhancing the stability of the motion plan | Static | Simulation |
| 23 | Path planning for mobile robots using bacterial potential field for avoiding static and dynamic obstacles | APF combined with bacterial evolutionary algorithm (BEA) for an enhanced flexible path planner | Static and dynamic | Simulation |
| 24 | Real time robot path planning method based on improved artificial potential field method | Modification and addition of attractive and repulsive forces | Static | Simulations and real time |
| 25 | Intelligent vehicle path planning based on improved artificial potential field algorithm | Used distance correction factor and global minimum regulatory factor. Also, proposed use of fuzzy algorithm for planning in local minimum location | Static | Simulation |
| 26 | Improved potential field method path planning based on genetic algorithm | Used genetic algorithm (GA) to optimize the combined potential field function of gravity and repulsion in the APF | Static | Simulation |
| 27 | Hierarchical motion planning and tracking for autonomous vehicles using global heuristic-based potential field and reinforcement learning-based predictive control | Used modified A-star algorithm and locally weighted regression smoothing incorporated with APF to generate the real-time optimal motion sequences | Static and dynamic | Simulation and real time |
| 28 | Path planning for robot based on chaotic artificial potential field method | Proposed combination of improved APF with chaotic optimization algorithm | Static | Simulation |
| 29 | An improved artificial potential field escape method with weight adjustment | Suggested modified APF approach by increasing the influence of gravity while reducing the influence of repulsion | Static and dynamic | Simulation |
| 30 | Development of modified path planning algorithm using artificial potential field (APF) based on PSO for factors optimization | Combined use particle swarm optimization (PSO) to find optimum of APF factor for shortest path. Also spline equation used for path smoothing | Static | Simulation |
| 31 | Research on path-planning algorithm integrating optimization a-star algorithm and artificial potential field method | Proposed a fusion-based path finding approach based on optimized A-star algorithm, the APF and the least squares method | Static and dynamic | Simulation |
| 32 | Development of A* algorithm for robot path planning based on modified probabilistic roadmap and artificial potential field | Combined use of APF with probabilistic roadmap (PRM) for enhancing the location of the nodes. While, A-star was used to find shortest path within the constructed map | Static | Simulation |